\documentclass[10pt,journal,compsoc]{IEEEtran}
\ifCLASSOPTIONcompsoc
  \usepackage[nocompress]{cite}
\else
  \usepackage{cite}
\fi

\usepackage{ragged2e}
\usepackage{epsfig}
\usepackage{booktabs}
\usepackage{subcaption}

\usepackage{algorithmic}
\usepackage{url}
\usepackage{enumerate}
\usepackage{bm}
\usepackage{multirow}
\renewcommand\figurename{Figure}
\usepackage{graphicx}
\usepackage[ruled]{algorithm2e}
\usepackage{amsmath}
\usepackage{CJK}

\SetAlFnt{\small}
\SetAlCapFnt{\small}
\SetAlCapNameFnt{\small}
\SetAlCapHSkip{0pt}
\IncMargin{-\parindent}
\DeclareMathSizes{14}{14}{9.8}{7}

\ifCLASSINFOpdf
\else

\fi
\begin{document}

%
\title{Detecting Adversarial Examples via Key-based Network}

\author{Pinlong~Zhao,~Zhouyu~Fu,~Ou~wu,~Qinghua~Hu,~and~Jun~Wang
\IEEEcompsocitemizethanks{\IEEEcompsocthanksitem Pinlong Zhao, Ou Wu, Qinghua Hu and Jun wang are with  Tianjin University, E-mail: \{pinlongzhao, wuou, huqinghua, jun.wang\}@tju.edu.cn

\IEEEcompsocthanksitem Zhouyu Fu is with Microsoft Research Asia, E-mail: zhofu@microsoft.com} 

}

\markboth{***********************, VOL. *, NO. *, * 2018}%
{Shell \MakeLowercase{\textit{et al.}}: Bare Demo of IEEEtran.cls for Computer Society Journals}

\IEEEtitleabstractindextext{%
\begin{abstract}
Though deep neural networks have achieved state-of-the-art performance in visual classification, recent studies have shown that they are all vulnerable to the attack of adversarial examples. Small and often imperceptible perturbations to the input images are sufficient to fool the most powerful deep neural networks. Various defense methods have been proposed to address this issue. However, they either require knowledge on the process of generating adversarial examples, or are not robust against new attacks specifically designed to penetrate the existing defense. In this work, we introduce key-based network, a new detection-based defense mechanism to distinguish adversarial examples from normal ones based on error correcting output codes, using the binary code vectors produced by multiple binary classifiers applied to randomly chosen label-sets as signatures to match normal images and reject adversarial examples. In contrast to existing defense methods, the proposed method does not require knowledge of the process for generating adversarial examples and can be applied to defend against different types of attacks. For the practical black-box and gray-box scenarios, where the attacker does not know the encoding scheme, we show empirically that key-based network can effectively detect adversarial examples generated by several state-of-the-art attacks.
\end{abstract}

\begin{IEEEkeywords}
adversarial example, neural network, defence, key-based network.
\end{IEEEkeywords}}

\maketitle

\IEEEdisplaynontitleabstractindextext

%
\IEEEpeerreviewmaketitle

\IEEEraisesectionheading{\section{Introduction}\label{sec:introduction}}

%
%
%
%
\IEEEPARstart{V}isual classification is particulary important in many applications \cite{Wang2564} and deep neural networks have achieved state-of-the-art performance \cite{he2016deep}. Despite their effectiveness for classification tasks, recent research showed that mainstream deep neural networks are all vulnerable to the attack of adversarial examples \cite{szegedy2013intriguing}, \cite{goodfellow2014explaining}, \cite{moosavi2016deepfool}, \cite{liu2016delving}. Specifically, very small and often imperceptible perturbations to the input images are sufficient to fool the most powerful deep neural network models and result in incorrect classification.

The vulnerability to adversarial inputs can be problematic and even prevent the application of deep learning methods in safety- and security-critical applications. The problem is particularly severe when human safety is involved.

To improve the robustness against adversarial examples for the deep neural networks, several defense methods have been proposed based on augmenting the training data with adversarial examples \cite{szegedy2013intriguing}, \cite{goodfellow2014explaining}, detecting the adversarial examples \cite{metzen2017detecting}, \cite{grosse2017statistical}, modifying either the input examples \cite{meng2017magnet} or the network architecture \cite{papernot2016distillation}. However, they either require knowledge on the process of generating adversarial examples \cite{szegedy2013intriguing}, \cite{goodfellow2014explaining}, \cite{metzen2017detecting}, or are not robust against new attacks specifically designed to penetrate the existing defense \cite{carlini2017towards} \cite{Carlini1711}.

In this paper, we propose key-based network, a novel defense method based on the detection of adversarial examples from normal input. Unlike existing detection-based defense methods \cite{metzen2017detecting}, \cite{grosse2017statistical} that leverage intermediate layer representations of the trained model for learning a binary classification model to distinguish between adversarial examples and normal images and thus require both adversarial and normal input data for the detection task, the proposed key-based network does not specifically train a binary classifier to detect adversarial examples and only requires normal examples in training. Inspired by the error-correcting output code framework \cite{dietterich1994solving}, key-based network trains multiple binary classifiers for randomly selected pairs of label subsets and encodes each class with a binary code vector based on the expected output for all individual classifiers. The binary encoding for each class can be viewed as a unique signature for the class and used to match normal examples and reject adversarial examples based on their output codes.

The main contributions in this paper are as follows: First, unlike previous detection methods in \cite{metzen2017detecting}, \cite{grosse2017statistical}, our proposed method requires normal examples only to train the detector and does not need to know the process of generating adversarial examples. Hence it can detect adversarial examples generated by different attacks. Second, one needs to know the exact encoding scheme to break our defense. However, the encoding schemes in the key-based network are randomly chosen and the number of potential schemes grows exponentially with increasing number of classes. This makes it very hard to devise new attacks to evade the detection of our method. Lastly, our method does not rely on the network architecture and can be used to protect a wide range of neural networks. Compared with the original network, key-based network suffers minimal accuracy loss on normal images, and can effectively detect adversarial examples.

\section{PRELIMINARIES}
In this section, we describe the preliminaries required to further elaborate on our defense method. These include definitions employed in this study, three mainstream adversarial attacks used in the experiments.
\subsection{Definitions}
\subsubsection{Examples}
Normal examples are all the examples from the original data set.

Adversarial examples are crafted by attackers using normal examples. They are perceptually close to normal examples but cause misclassification in the classifier.
\subsubsection{Networks}
Original networks are trained with all the normal examples in the training set.

Key-based networks are also trained with the normal examples and used to detect adversarial examples.
Key-based networks share the same structure as the corresponding original networks and only differ in the logits layer and softmax layer, as shown in Figure 1.


\subsection{Generating adversarial examples}
\label{ssec:adversarial-attacks}
\subsubsection{Fast gradient sign method (FGSM)}
Given a normal image $x$, FGSM \cite{goodfellow2014explaining} looks for a similar image $x'$ in the $L_\infty$ neighborhood of x that fools the classifier. Let $\theta$ be the parameters of a model, $x$ be the input example, $y$ be the target label and $Loss(\theta, x, y)$ be the cost associated with assigning label $y$ to example $x$ for the trained model. The adversarial example $x'$ can be obtained by maximizing the loss w.r.t. the input image $x$ in the local neighborhood of $x$. This leads to the following update rule:
\begin{equation}
x'=Proj_{B_r(x)}(x+\epsilon sign(\nabla_x Loss(\theta,x,y) ))
\end{equation}
where $\epsilon$ is the step size, and $Proj$ is the projection operator that maps the perturbed image to the $L_\infty$  neighborhood of input $x$ defined by $B_r(x)=\{{y:  \Vert{y-x} \Vert_{\infty}{\le}r}\}$. Note an approximate gradient is sufficient for adversarial attack, hence the sign of the gradient is used in the above equation due to efficiency concerns.
\subsubsection{Deepfool}
FGSM is a one-step attack method. It can be enhanced with iterative attack techniques by repeatedly applying one-step attacks. Moosavi-Dezfooli et al. \cite{moosavi2016deepfool} proposed such an attack algorithm namely Deepfool through the iterative linearization of the classifier to generate minimal perturbations that are sufficient to change classification labels. The basic idea is to find the closest decision boundary from a normal image $x$ in the image space, and then to cross that boundary to fool the classifier. Since it is hard to solve this problem directly in the high-dimensional and highly non-linear space in neural networks, it instead solves this problem iteratively with linearized approximations.

\begin{figure}[htbp]
\begin{center}
  \hspace{0in}\includegraphics[width= 0.45\textwidth, height = 135pt]{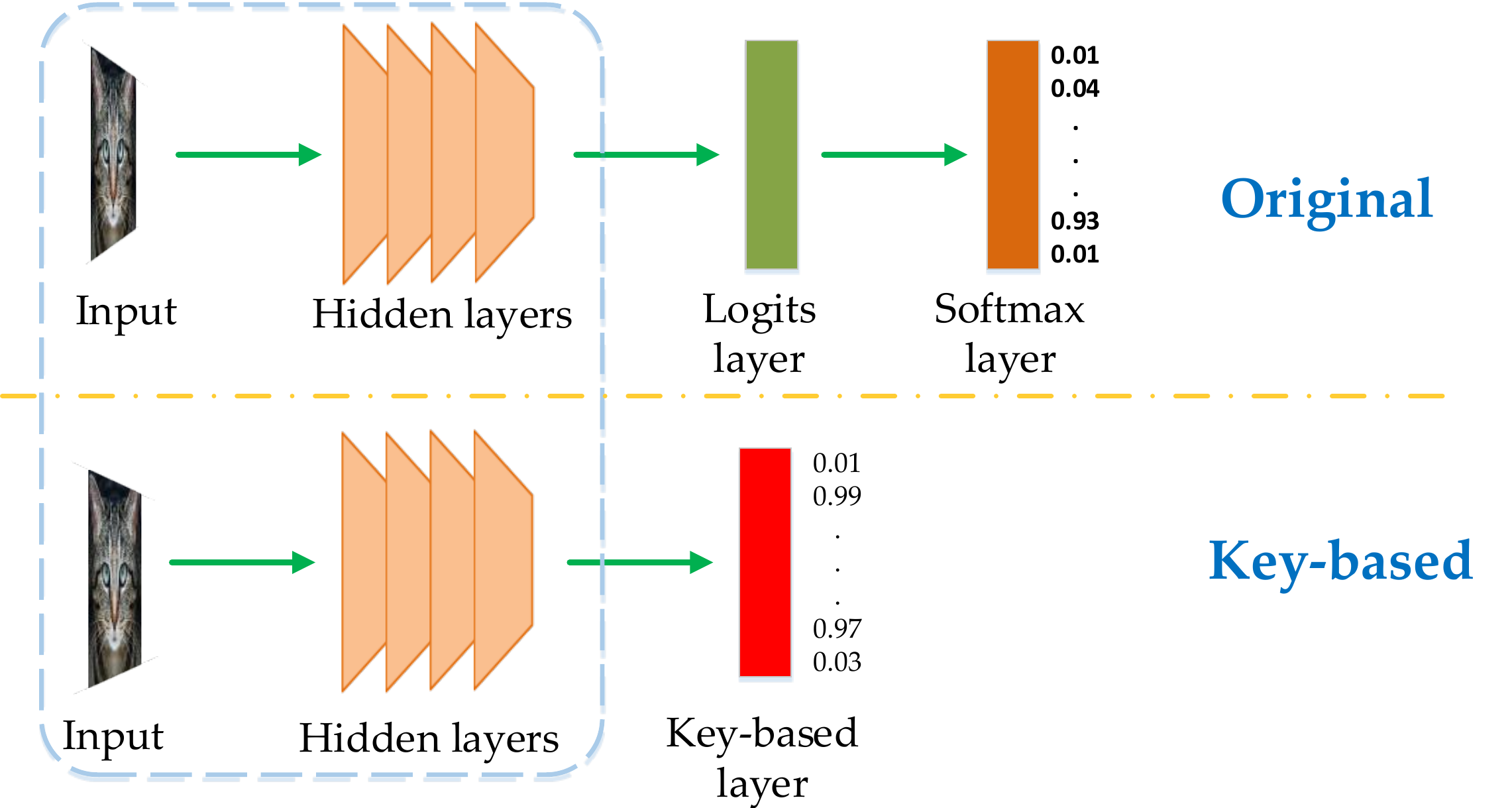}\\
  \vspace{-0.05in}\caption{ The original and the key-based network: The upper part of the figure is the original network, and the lower part is the key-based network. The dashed boxes indicate that they have the same structure.}\vspace{-0.15in}
\end{center}
\end{figure}

At each iteration, the classifier is linearized arround the intermediate $x'$, and the optimal update direction is obtained on this linearized model. Then $x$ will update a small step $\alpha$ in this direction. Finally an adversarial example will be obtained by repeatedly linearizing the update process until $x'$ crosses the decision boundary. The $L_\infty$ version of the deep fools attack is used in this paper.
\subsubsection{Carlini's attack}
Carlini and Wagner recently introduced a powerful attack that makes the perturbations quasi-imperceptible by restricting their $L_2$, $L_\infty$ and $L_0$ norms \cite{carlini2017towards}. For a fixed input image $x$, the attack looks for a perturbation $\delta$ that is small in length,satisfies the box constraints to be a valid image, and fools the classifier at the same time.

In Carlini's attack, there is a hyper-parameter called confidence. Higher confidence produces adversarial examples with larger distortion, but more strongly classified as adversarial.

For this paper, we use the untargeted $L_2$ version of Carlini's attack.
\subsection{Adversarial attacking scenarios}
All three attacks discussed in the previous subsection assumes the standard white-box attacking scenario, where the network structure and weights are known to the attacker. In contrast, for black-box attacks, the attacker has no knowledge on either the structure or the weights of the network to be attacked. This however gives no guarantee of security. Since different deep neural networks used in visual classification produce highly similar feature maps, it is always possible to train a surrogate network with the same dataset to simulate the original network. One can then launch effective adversarial attacks on the surrogate network instead \cite{liu2016delving}\cite{papernot2017practical}.

We focus on defense against gray-box attacks in this paper. In a gray-back attack, the attacker knows both the original network and the defense algorithm. Only the parameters of the defense model are hidden from the attacker. This is also a standard setting assumed in many security systems and applications \cite{pfleeger2002security}.

\section{KEY-BASED NETWORK}
In this section, we introduce key-based network. First, we discuss the main design of key-based network motivated by core ideas from cryptography. We then review the error-correcting output codes scheme that motives the proposed encoding scheme. Finally, we discuss the robustness of key-based network and highlight its differences from the standard gradient obfuscation based methods for defense.
\subsection{Design of Key-Based Network}
A major issue with existing defenses is their robustness, mainly due to the lack of enough random factors in their design, making it possible for the attackers to infer the structure and parameters of the defense model and launch new attacks to penetrate existing defenses using surrogate networks \cite{carlini2017magnet}.

To enable effective defenses against adversarial examples in the gray-box scenario, we need to increase the space of design parameters for the defense model and choose a random design parameter to achieve improved robustness. This is akin to the design of a cryptographical system, where we can view the defense model as the encryption algorithm, and the design parameters as the secret keys. The algorithm itself is not a secret. Only the keys are held secret, but there are exponential number of possible key values to choose from. Without knowing the exact key, there is no way to break the encryption algorithm \cite{dietterich1994solving}.

This motivates the key-based network for defending against adversarial examples. Since an attacker needs to know the label of an input example so as to maximize the loss of the true label, we treat labels as the equivalent plaintext to be protected. To this end, we devise an encryption scheme to encode each output label with a corresponding code vector. Formally, given an image $x$ with ground truth $y$, we have:
\begin{equation}
\begin{array}{l}
c=e_\theta(y){ ,\,\,\,\,\quad  }\forall{c}\in{C}\\
y=e_\theta^{-1}(c){ ,\quad  }\forall{c}\in{C}
\end{array}
\end{equation}
where $c$ is the code vector for class $y$, $e_\theta$ is the encryption function that maps label $y$ to code vector $c$, and $e_\theta^{-1}$ is the decryption function that maps $c$ to $y$,  $\theta$ is the parameter of the encryption function.

Since the encryption scheme defined above creates a one-to-one correspondence between labels and the code vectors they map to, it is possible to learn a network to fit input $x$ to the code vector $c$. In contrast, the original network learns a model to directly predict label $y$. Figure 1 shows the architectures of the original network and the key-based network. The hidden layers of key-based network have the same structure as the original network. The modification is in the output layer and the loss function layer (for training only), where the original network maps the output feature vector to logits of m classes, and key-based network maps the output feature vector to the corresponding code vector. We can treat the original network as the special case where the encryption function is a one-hot encoding function that maps a discrete label to a one-hot vector.

For classification network, the following cross entropy is usually employed as the loss function:
\begin{equation}
L = -\frac{1}{n}\sum_{i=1}^n\sum_{j=1}^m(y_j(x_i){\,\,}{\mathop{\rm log}\nolimits}f_j(x_i))
\end{equation}
where $n$ is the size of the batch size.  $y$ is an indicator vector that the only non-zero element corresponds to the correct class, $y_j$ is $j$th component of $y$.

For key-based layer training, we employ the squared loss to measure the discrepancy between the output and code values:
\begin{equation}
L = -\frac{1}{n}{\,\,}{\frac{1}{t}}{\,}\sum_{i=1}^n\sum_{j=1}^t{(f_j(x_i)-c_j^{-1}(y_i))^2}
\end{equation}
where  $c_j^{-1}(y_i)$ is $j$th component of $c^{-1}(y_i)$.

With key-based network, we can view code vector $c$ as the signature for label $y$. This leads to a natural strategy to detect adversarial examples, by verifying whether the code vector computed from the input example matches the signature of any class within certain precision level. If the output code vector does not match any signature, the corresponding input is treated as the adversarial example.
\begin{figure}[htbp]
\begin{center}
  \hspace{0in}\includegraphics[width= 0.40\textwidth, height = 120pt]{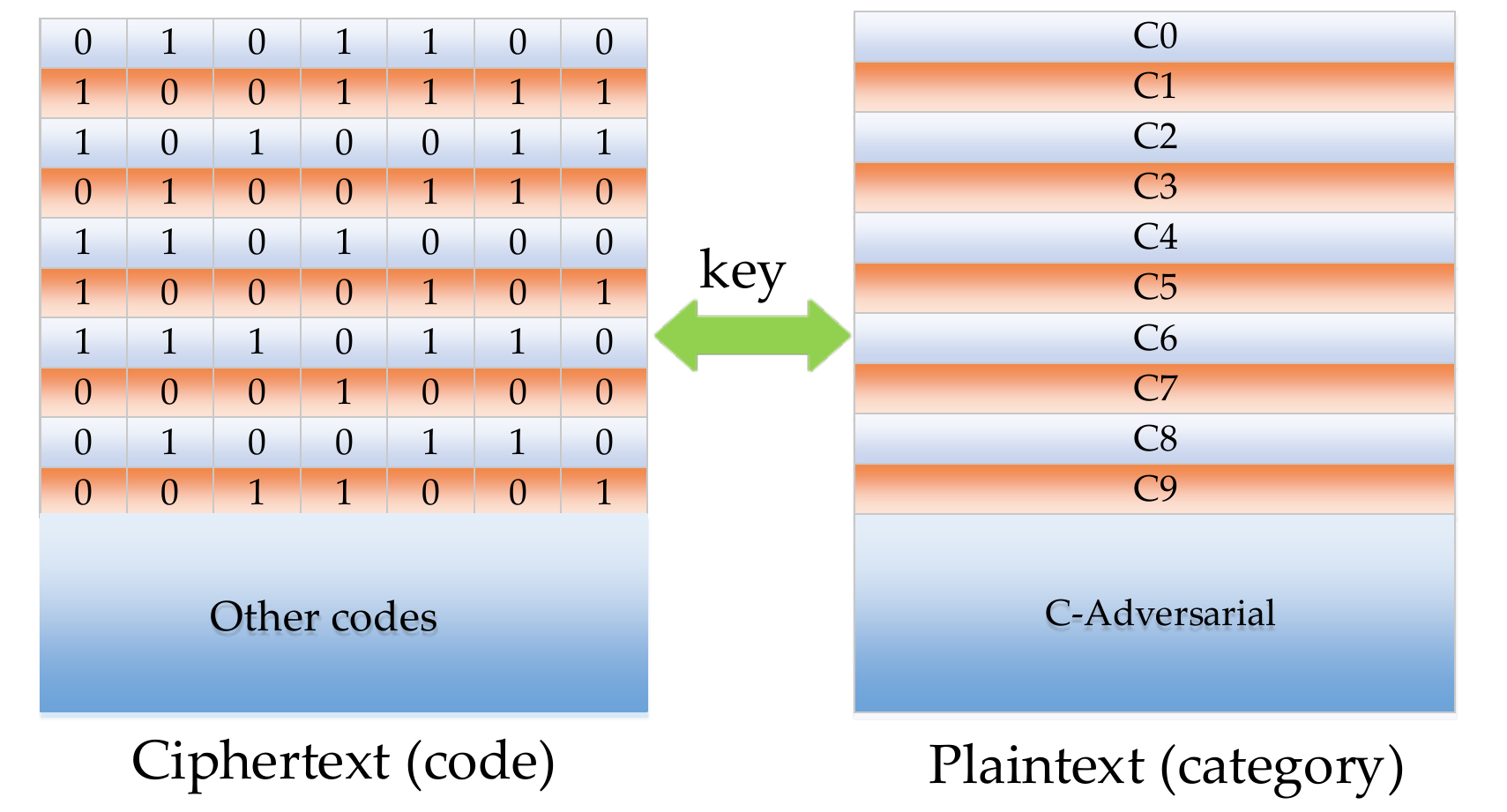}\\
  \vspace{-0.05in}\caption{ There is a one-to-one match between each code and each category. The left table is a code table, and each line represents a different code. In particular, this is a 7-bit code.
  There are $2^7$ different codes.
  The table on the right indicates that there are 11 cases: 10 different classes, and the adversarial examples. Through key, the plaintext (code) can be transformed into a plaintext (the corresponding category).}\vspace{-0.15in}
\end{center}
\end{figure}

\subsection{Encoding Scheme of Key-Based Network}
We used a lookup table to implement the encryption function in Equation (1), by mapping each label directly to a code vector. The encoding scheme is essential for the security of the detection algorithm. An example is shown in Figure 2, where the lookup table is a matrix composed by stacking the code vectors for each label class in rows. Each column defines a binary classification problem by treating labels corresponding to value of 1 as the positive class and labels with value of 0 as the negative class. This connects key-based network to the error correcting output codes (ECOC) framework \cite{dietterich1994solving} for multi-class classification problem. ECOC includes the one-vs-all scheme as a special case, which is essentially the case of an identity matrix for the lookup table and has been exclusively used in visual classification.

To make the encoding scheme elusive for the attacker to break it, we use randomly generated code vector for each class. For a $m$-class classification problem, there is a total of $2^{m-1}-1$ different assignment of binary problems for the encoding scheme as equivalent to the columns of the lookup table, excluding all 1 and all 0 assignments and mirrored assignments obtained by swapping the labels. Hence the complexity of testing for the right encoding scheme to break the defense algorithm grows exponentially with the number of classes, making it infeasible for the attacker to penetrate the system.

Note though we describe and employ binary encoding scheme here, key-based network defines a broad frame-work and can virtually incorporate with different types of encodings, such as $n$-way ($n>2$) or $k$-bit ($k>1$) encoding for each entry in the lookup table. This leads to multi-class classification for each column and is likely to boost the strength of defense albeit at the cost of complexity.

\begin{figure}[htbp]
\begin{center}
  \hspace{0in}\includegraphics[width= 0.40\textwidth, height = 104pt]{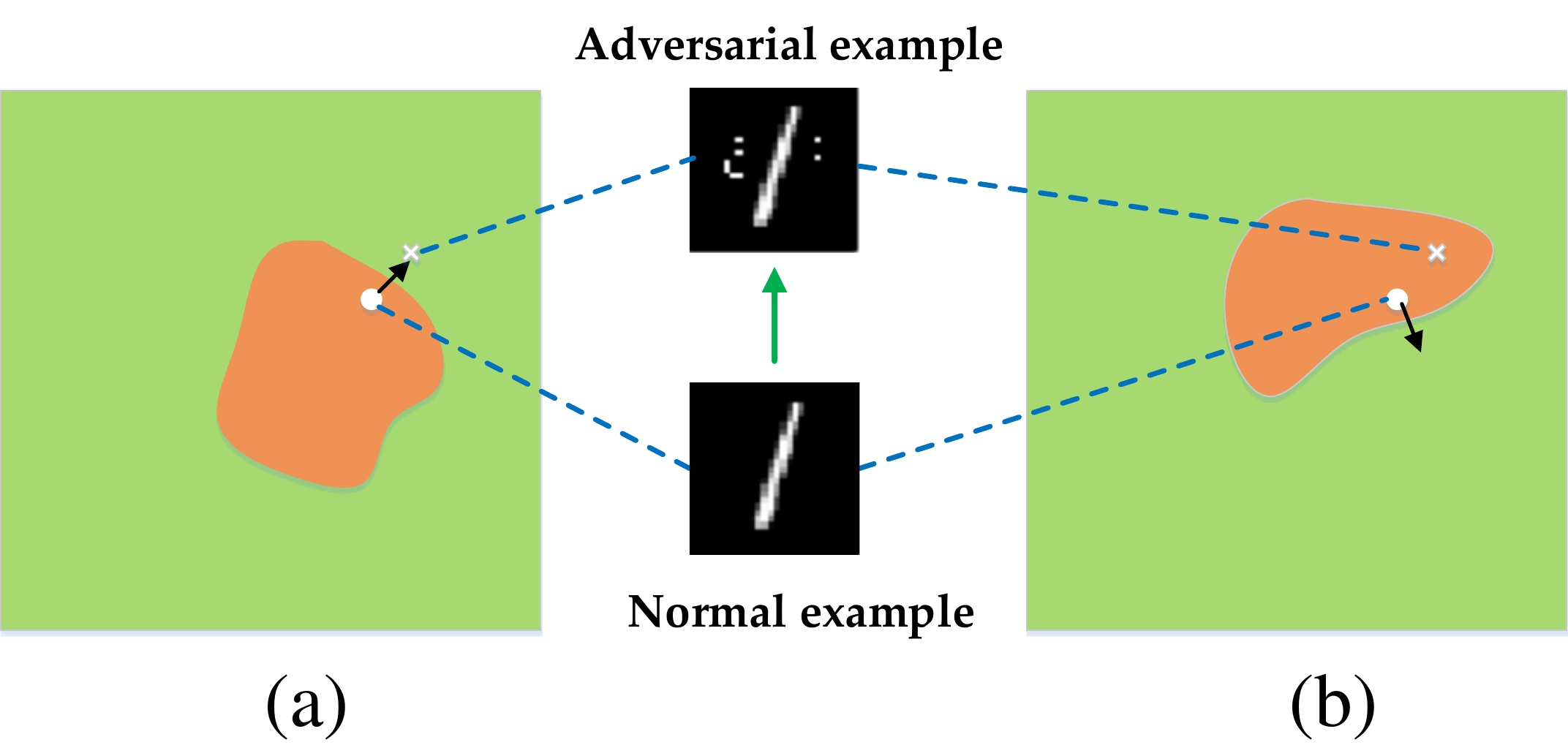}\\
  \vspace{-0.05in}\caption{ Illustration of how gradient masking method and key-based network method are different in a 2-D sample space. We represent the boundary by orange curve. The arrow indicates sensitive direction.
  }\vspace{-0.15in}
\end{center}
\end{figure}

\subsection{Robustness of Key-Based Network}
Our method is based on attackers' ignorance of key-based layer's parameters and key. First, the network parameters of key-based layer are not known by attackers, as long as we do not release its internal states. Attackers know the output classes of network, the network's structure and parameters of network except for key-based layer. Second, an attacker can hardly guess the exact code when he doesn't know the key.

Although our method has the same characteristics as gradient masking: the attacker can not access to the gradient, our method is quite different from gradient masking. Figure 3 shows a simple example to illustrate the difference between key-based network method and gradient masking method. For an input $x$, the attacker's purpose is to find a small enough perturbation $\epsilon$, so that $x$ and $x+\epsilon$ are on different sides of the boundary (changing the classification result). The direction which from x to $x+\epsilon$ is called sensitive directions. Inputs that make a small change in this direction can lead to misclassification. Gradient masking does not expose these directions to defend adversarial examples, but these directions still exist. These directions can be discovered by black-box attack method \cite{papernot2017practical}. For the key-based network, due to the change of the input to output mapping, the boundary also changes, resulting in a change of sensitive direction. As shown in Figure 3, attackers can not know the exact sensitive direction. Therefore, this method can effectively defend against adversarial examples.



\section{EMPIRICAL EVALUATION}
In this section, we empirically evaluate key-based network, using the MNIST \cite{lecun1998mnist} dataset for hand-written digit recognition. First, we show that adversarial examples can be detected with our proposed method for all three popular attacks discussed in Section \ref{ssec:adversarial-attacks}. We then examine the effect of code lenghts on the detection performance. Finally, we empirically evaluate why key-based network is effective.

%
%
%
%
%
%

\subsection{Experiment Setup}
\subsubsection{Dataset Description}
The MNIST dataset comprises a training set of 60,000 grayscale images of handwritten digits, and a test set of 10,000 images. Each image represents one of the $10$ digits from $0$ to $9$ with a resolution of  28*28 pixels.
\subsubsection{Evaluation Criterion}
We employ two metrics, namely predictive accuracy and detection rate, to evaluate the performance of the proposed key based network. Predictive accuracy is defined for normal images as the ratio of correctly classified normal examples to all normal examples. For key-based network, an example is correctly classified if and only if the output code matches exactly the signature of the ground truth class. Predictive accuracy measures how well the modified network can deal with normal examples. Detection rate is defined for adversarial images as the ratio of adversarial examples correctly detected by key based network to the total number of adversarial examples generated. Ideally, we would expect the proposed method to achieve very high detection rate without losing much predictive accuracy as compared to the deep neural network trained with standard method.
\begin{figure}[htbp]
\begin{center}
  \hspace{0in}\includegraphics[width= 0.49\textwidth, height = 90pt]{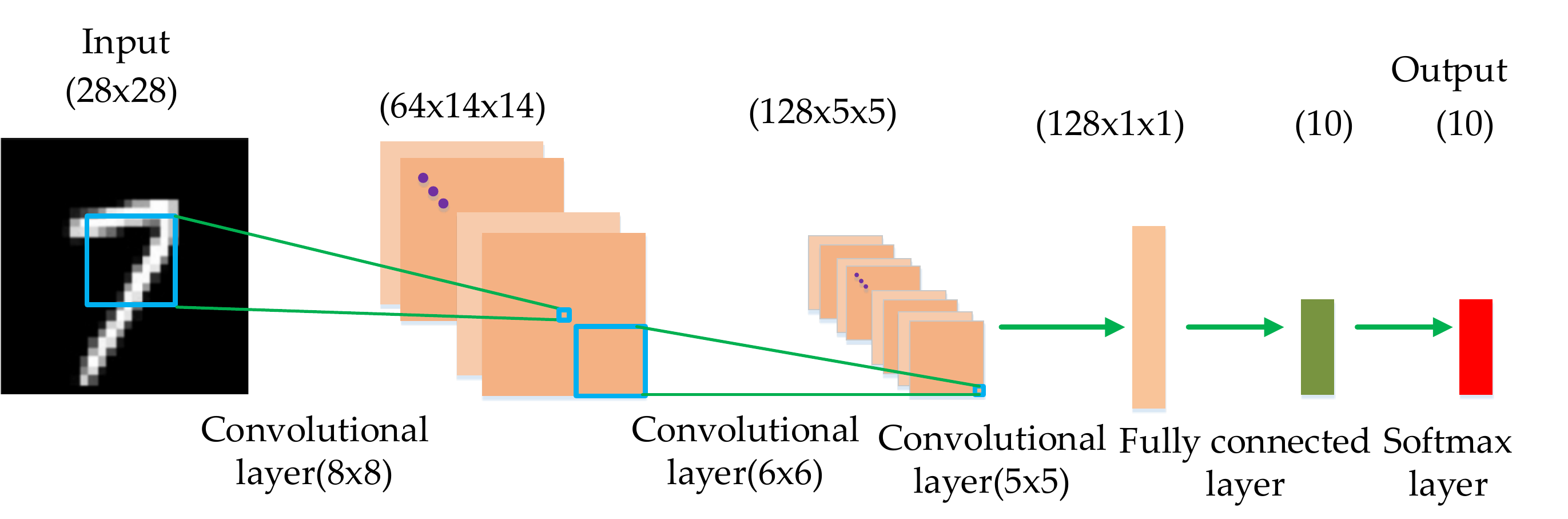}\\
  \vspace{-0.05in}\caption{ The baseline model.}\vspace{-0.15in}
	\label{fig:baseline_network}
\end{center}
\end{figure}
\subsubsection{Architecture of Baseline Network}
We used the Convolutional Neural Network (CNN) in Figure 4 as the baseline model for the classification task. It contains three convolutional layers with RELU activation function for each layer, a fully connected layer, and a softmax layer as an output. We will evaluate a variety of other models in Section \ref{ssec:empirical-analysis} to demonstrate the effectiveness of key based network.

\begin{table}
\renewcommand\arraystretch{1.18}
\setlength{\abovecaptionskip}{-10pt}
\setlength{\belowcaptionskip}{10pt}
\begin{center}
\caption{}\label{table:performance}
\begin{subtable}{2in}
\centering
\caption{Accuracy on normal examples}\label{table:performance:accuracy}
 \begin{tabular}{lclcc}
 \toprule
  &Baseline&& Knet&  \\
 \midrule
 &99.25\% & & 98.62\%& \\
 \bottomrule
 \end{tabular}
 \end{subtable}
\begin{subtable}[t]{3in}
\centering
\setlength{\belowcaptionskip}{10pt}
 \caption{Detection rate on adversarial examples}\label{table:performance:detection_rate}
 \begin{tabular}{lclc}
 \toprule
  Attack & Parameter & Fooling Rate & Detection Rate\\
 \midrule
 FGSM & $\epsilon$=0.2 & 49.60\% & 96.26\%\\
 FGSM & $\epsilon$=0.3 & 88.52\% & 99.31\%\\
 DeepFool &  & 99.35\% & 97.83\%\\
 Carlini & $\kappa$=0.0 & 100.0\% & 95.66\%\\
 Carlini & $\kappa$=0.1 & 100.0\% & 94.86\%\\
 Carlini & $\kappa$=0.2 & 100.0\% & 94.14\%\\
 \bottomrule
 \end{tabular}
 \end{subtable}
 \end{center}
\end{table}
\subsection{Adverarial Example Detection}

We first show that key based network can effectively detect adversarial examples generated by three popular attacks, namely FGSM, DeepFool and Carlini's attack.
We first use all training examples in the MNIST dataset to train the baseline network shown in Figure 4. Then the output layer and the softmax loss layer of the trained baseline network were replaced with key-based layer and the squared loss layer to retrain the model so as to obtain the key-based network for the detection of adversarial examples.

In the testing process, we first gain the  predictive accuracies  on 10,000 normal test examples for both the baseline network and key-based network. We then generate adversarial examples from test images correctly classified by the baseline network using each of the three above-mentioned attacks and verify if the adversarial examples generated by these different attacks can be correctly detected by the key-based network. An adversarial example is detected correctly if either it can be detected as an adversarial example or it can be classified correctly by the modified network.

The results are reported in Table \ref{table:performance}, where Table \ref{table:performance:accuracy} shows the accuracies of the baseline model and key-based network on normal test examples, and Table \ref{table:performance:detection_rate} shows the detection rate of key-based network on adversarial test examples generated by different attacks, along with the fooling rates of the generated adversarial examples for the baseline model. The results indicate that the key-based network achieves comparable predictive performance with the baseline network, but is much more robust to adversarial examples and can successfully defend against all adversarial attacks we tested. With only $0.6\%$ drop in accuracy, the proposed method achieves a detection rate of over $94\%$ for Carlini's attack, which is able to achieve $100\%$ fooling rate for the baseline model. The detection rate achieve for  FGSM and Deepfool is even high and can reach $99\%$ for FGSM. On the other hand, even for the weakest FGSM attack, the baseline model can still be fooled by half of the adversarial examples.

\subsection{Effect of Code Lengths on Detection Performance}
In the next experiment, we examine how the length of the output code affects the performance of key-based network. We train and test our network with code lengths of 5, 10, 20, 30, 40, 50 and 60, respectively. The results are shown in Figure 5 and Figure 6.

Figure 5 shows the changes in predictive accuracies for normal examples with different code lengths. The longer the code, the lower the accuracy is, though the reduction in accuracy is quite small. With the code length changing from 5 to 60, the predictive accuracy drops from 98.7\% to 97.7\%. Figure 6 shows the changes in detection rates for adversarial examples. For all the three attacks, the detection rate improves with increasing length of codes. When the length of code is larger than 40, the detection rate exceeds 97\% for all three attacks. The reason is the larger the code length, the harder it is to decipher the exact code, and hence the more secure the key-based network is.
\begin{figure}[htbp]
\begin{center}
  \hspace{0in}\includegraphics[width= 0.49\textwidth, height = 100pt]{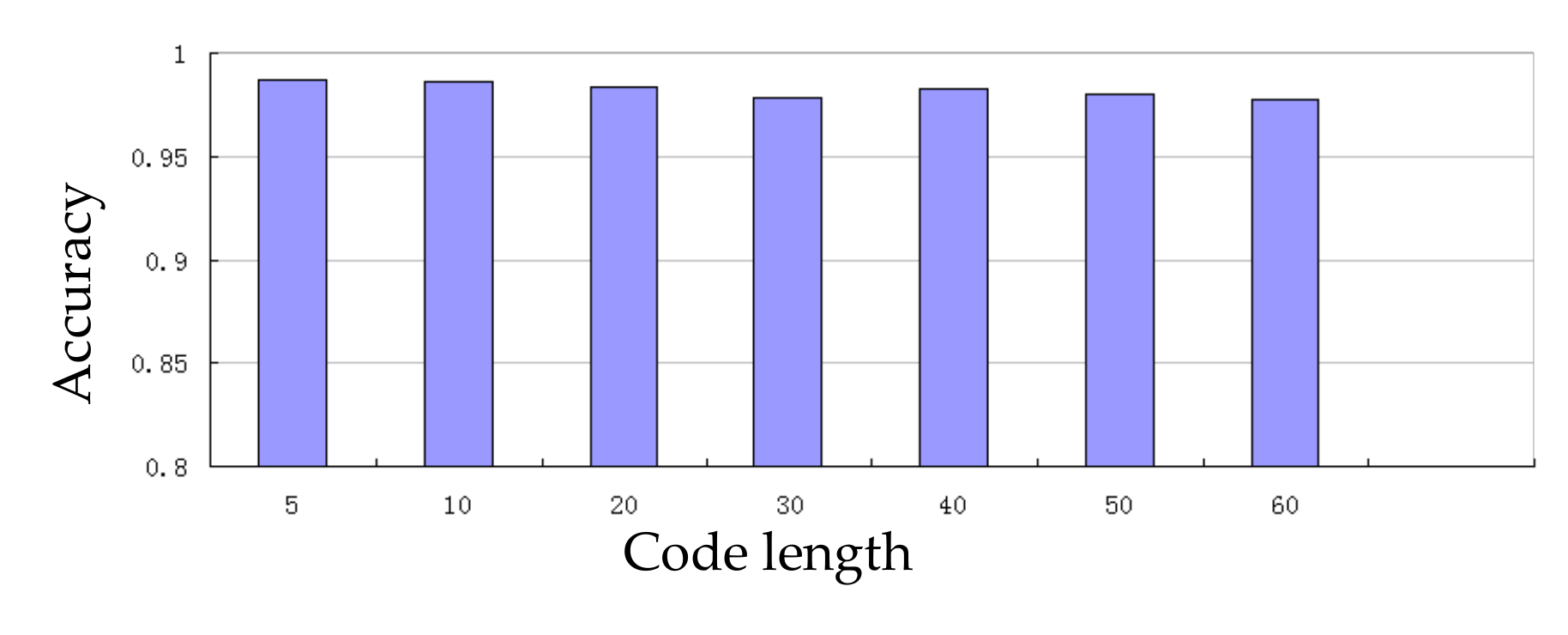}\\
  \vspace{-0.05in}\caption{The accuracy on different code lengths.}\vspace{-0.15in}
\end{center}
\end{figure}
\begin{figure}[htbp]
\begin{center}
  \hspace{0in}\includegraphics[width= 0.49\textwidth, height = 160pt]{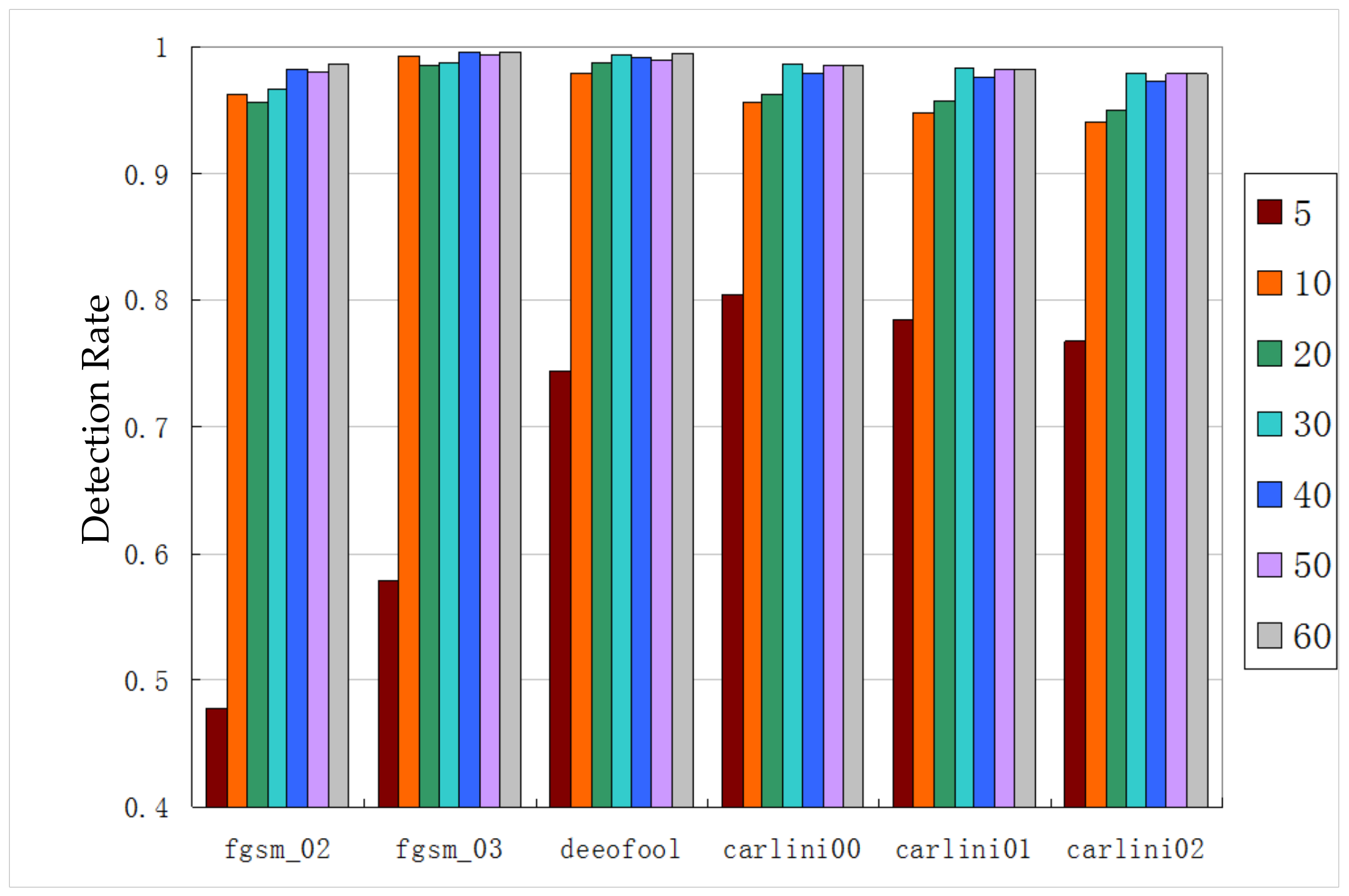}\\
  \vspace{-0.05in}\caption{ Detection rate of different code lengths.}\vspace{-0.15in}
\end{center}
\end{figure}

Overall, it can be seen that the performance of key-based network is not sensitive to the choice of code lengths provided a sufficiently long code is used. The performance is quite stable over a wide range of code lengths with no evidence of sudden drop in accuracy or boost in detection rate with increasing code lengths.

\begin{table}
\renewcommand\arraystretch{1.18}
\setlength{\abovecaptionskip}{00pt}
\setlength{\belowcaptionskip}{00pt}
\caption{Network with different architectures}\label{table:baseline_networks}
\begin{center}
\begin{subtable}{3in}
\centering
\caption{Baseline2}\label{table:baseline2}
 \begin{tabular}{lclcc}
 \toprule
  &size &stride & padding&  \\
 \midrule
  Conv.ReLU&8*8 (64)&2*2 & same& 14*14*64  \\
  Conv.ReLU&6*6 (128)&2*2 & valid& 5*5*128  \\
  Conv.ReLU&5*5 (128)&1*1 & valid& 1*1*128  \\
  FullyConnect&10& & & 100  \\
  FullyConnect&10& & & 10  \\
  Softmax&10& & & 10  \\
 \bottomrule
 \end{tabular}
 \end{subtable}
\begin{subtable}[t]{3in}
\centering
\setlength{\belowcaptionskip}{10pt}
 \caption{Baseline3}\label{table:baseline3}
 \begin{tabular}{lclcc}
 \toprule
  &size &stride & padding&   \\
 \midrule
  Conv.ReLU&3*3 (64)&2*2 & same& 14*14*64  \\
  Conv.ReLU&3*3 (128)&2*2 & valid& 6*6*128  \\
  Conv.ReLU&5*5 (128)&1*1 & valid& 1*1*128  \\
  FullyConnect&10& & & 100  \\
  FullyConnect&10& & & 200  \\
  FullyConnect&10& & & 10  \\
  Softmax&10& & & 10  \\
 \bottomrule
 \end{tabular}
 \end{subtable}
 \begin{subtable}[t]{3in}
\centering
\setlength{\belowcaptionskip}{10pt}
 \caption{Baseline4}\label{table:baseline4}
 \begin{tabular}{lclcc}
 \toprule
  &size &stride & padding&   \\
 \midrule
  Conv.ReLU&3*3 (128)&2*2 & valid& 13*13*128  \\
  Conv.ReLU&5*5 (128)&2*2 & valid& 6*6*128  \\
  FullyConnect&10& & & 100  \\
  FullyConnect&10& & & 10  \\
  Softmax&10& & & 10  \\
 \bottomrule
 \end{tabular}
 \end{subtable}
 \end{center}
\end{table}

\subsection{Empirically analysis on the Effectiveness}
\label{ssec:empirical-analysis}
In the last experiment, we provide an empirical analysis on the effectiveness of the proposed key-based network by examining its performance overlap with the baseline model. To enable a more thorough comparison,
in addition to the baseline network shown in Figure \ref{fig:baseline_network},  we have trained three other baseline networks on the same training set of 60,000 images with different architectures, namely  baseline2, baseline3, baseline4.
Table \ref{table:baseline_networks} shows the architectures of these networks.
We first apply the trained networks, include the four baseline models and key-based network, to 10,000 test examples with randomly generated pixel values in the range of $[0,1]$.
The number of times that the output labels produced by the baseline2-4 models and key-based network overlap with the output labels of the baseline network are calculated and the result is shown in Table \ref{table:analysis:random_input}.
We then repeat the same test for normal test images with random noises.  100 images were randomly sampled from 10,000 test images. For each image, we apply 100 different random noise patterns by independently adding
a pixelwise uniform random noise in the range $[-0.5, 0.5]$ and clipping the result to $[0, 1]$ range. This also leads to 10,000 test examples for the second test. The result on overlapping prediction labels
is shown in Table \ref{table:analysis:random_noise}.

From the results in Table \ref{table:analysis}, it can be seen that though the architectures of baseline2, baseline3, baseline4 networks are all different from the baseline network, their prediction results have a very high overlapping ratio for random inputs and images imposed with random noises. This reveals highly correlated input-output mappings learned by these baseline models despite their differences in the architecture, and suggests the great chance on the success of tranfer attacks to break the baseline system.
On the other hand, the overlapping in prediction labels is very low between key-based network and the baseline model, which suggests potentially different mappings learned by the two networks.
Therefore, the adversarial examples generated by other networks can hardly be transferred to key-based network.

\begin{table}
\renewcommand\arraystretch{1.18}
\setlength{\abovecaptionskip}{-10pt}
\setlength{\belowcaptionskip}{10pt}
\begin{center}
\caption{}\label{table:analysis}
\begin{subtable}[t]{3in}
\centering
\caption{Random inputs}\label{table:analysis:random_input}
 \begin{tabular}{lclc}
 \toprule
  Baseline2&Baseline3 &Baseline4 & Knet  \\
 \midrule
  4826&5748&3763 & 0 \\
 \bottomrule
 \end{tabular}
 \end{subtable}
\begin{subtable}[t]{3in}
\centering
\setlength{\belowcaptionskip}{10pt}
 \caption{Normal examples with random noise}\label{table:analysis:random_noise}
 \begin{tabular}{lclc}
 \toprule
  Baseline2&Baseline3 &Baseline4 & Knet  \\
 \midrule
  9807&9663&9373 & 1948 \\
 \bottomrule
 \end{tabular}
 \end{subtable}
 \end{center}
\end{table}

\section{Conclusion}
Detecting adversarial examples is one of defenses against adversarial examples. While this method does not directly allow classifying adversarial examples correctly, it allows mitigating adversarial attacks against machine learning systems by resorting to fallback solutions. However, prior work in detecting adversarial examples requires knowledge of adversarial examples. Thus, the defense is specific to the process for generating those adversarial examples. In this paper, we proposed key-based network, a framework for detecting adversarial examples. Key-based network learns to detect adversarial examples without requiring knowledge of adversarial examples, which leading to better generalization in different attacks. Experiments show that key-based network can detect the popular attacks effectively, while suffers minimal accuracy loss on normal examples. Future work should investigate the impact of key-based network on other DNN models and adversarial examples crafting algorithms.


%

\appendices

\ifCLASSOPTIONcompsoc
  \section*{Acknowledgments}
\else
  \section*{Acknowledgment}
\fi
The authors wish to thank Shunzhi Jiang for the the revision of this paper.
\ifCLASSOPTIONcaptionsoff
  \newpage
\fi



\bibliographystyle{IEEETran}
\bibliography{myk_net}
%




\end{document}